\documentclass{article}
\usepackage{spconf,amsmath,graphicx}
\usepackage[list=true,labelformat=brace, position=top]{subcaption}
\usepackage{multirow}
\usepackage[font=small]{caption}	
\usepackage{setspace}
\usepackage{tikz}
\usepackage{hyperref}

\newcommand\QP{\mathit{QP}}
\def\inputImage{\boldsymbol{I}}

\def\salientCTUs{{H}}
\def\detections{\boldsymbol{D}}
\def\detectionsYOLO{S}

\title{Saliency-Driven Versatile Video Coding for Neural Object Detection}
\name{Kristian Fischer, Felix Fleckenstein, Christian Herglotz, Andr\'e Kaup \thanks{The authors gratefully acknowledge that this work has been supported by the Deutsche Forschungsgemeinschaft (DFG) under contract number KA 926/10-1.}}
\address{Multimedia Communications and Signal Processing\\Friedrich-Alexander-Universit\"at Erlangen-N\"urnberg (FAU), Cauerstr. 7, 91058 Erlangen, Germany}
\newcommand\copyrighttext{%
	\footnotesize \textcopyright 2020 IEEE. Personal use of this material is permitted.
	Permission from IEEE must be obtained for all other uses, in any current or future 
	media, including reprinting/republishing this material for advertising or promotional 
	purposes, creating new collective works, for resale or redistribution to servers or 
	lists, or reuse of any copyrighted component of this work in other works. 
	DOI: \href{https://doi.org/10.1109/ICASSP39728.2021.9415048}{10.1109/ICASSP39728.2021.9415048} }
\newcommand\copyrightnoticeOwn{%
	\begin{tikzpicture}[remember picture,overlay]
		\node[anchor=north,yshift=-10pt] at (current page.north) {\fbox{\parbox{\dimexpr\textwidth-\fboxsep-\fboxrule\relax}{\copyrighttext}}};
	\end{tikzpicture}%
}

\begin{document}
%
\maketitle
\copyrightnoticeOwn
\vspace{-3mm}
\begin{abstract}
	Saliency-driven image and video coding for humans has gained importance in the recent past.
	In this paper, we propose such a saliency-driven coding framework for the video coding for machines task using the latest video coding standard Versatile Video Coding (VVC).
	To determine the salient regions before encoding, we employ the real-time-capable object detection network You Only Look Once~(YOLO) in combination with a novel decision criterion. To measure the coding quality for a machine, the state-of-the-art object segmentation network Mask R-CNN was applied to the decoded frame. From extensive simulations we find that, compared to the reference VVC with a constant quality, up to 29\,\% of bitrate can be saved with the same detection accuracy at the decoder side by applying the proposed saliency-driven framework. Besides, we compare YOLO against other, more traditional saliency detection methods.
\end{abstract}
\vspace{-2mm}
\section{Introduction}
In conventional video coding, each frame is commonly coded with a constant quantization parameter~(QP) depending on a user-selected base QP. But as shown in~\cite{helmrich2019}, a block-wise QP adaptation~(QPA) based on the human visual system is able to improve the subjective coding quality for humans as final observer, when coding tree units~(CTUs) covering non-salient areas are transmitted with a lower visual quality. This QPA functionality is included in the new video coding standard Versatile Video Coding~(VVC)~\cite{bossen2019} and its reference software VVC Test Model~(VTM)~\cite{chen2019vtm6}.

However, more and more of today's multimedia data traffic is not consumed by humans. Instead, computer vision algorithms analyze the data to solve different tasks, e.g., in the field of surveillance or autonomous driving. Thus, MPEG introduced an ad-hoc group on this so-called video coding for machines~(VCM) task to optimize video codecs for machine-to-machine communication scenarios~\cite{zhang2019}.

\begin{figure}[t]{}
	\centering
	\includegraphics[width=0.45\textwidth]{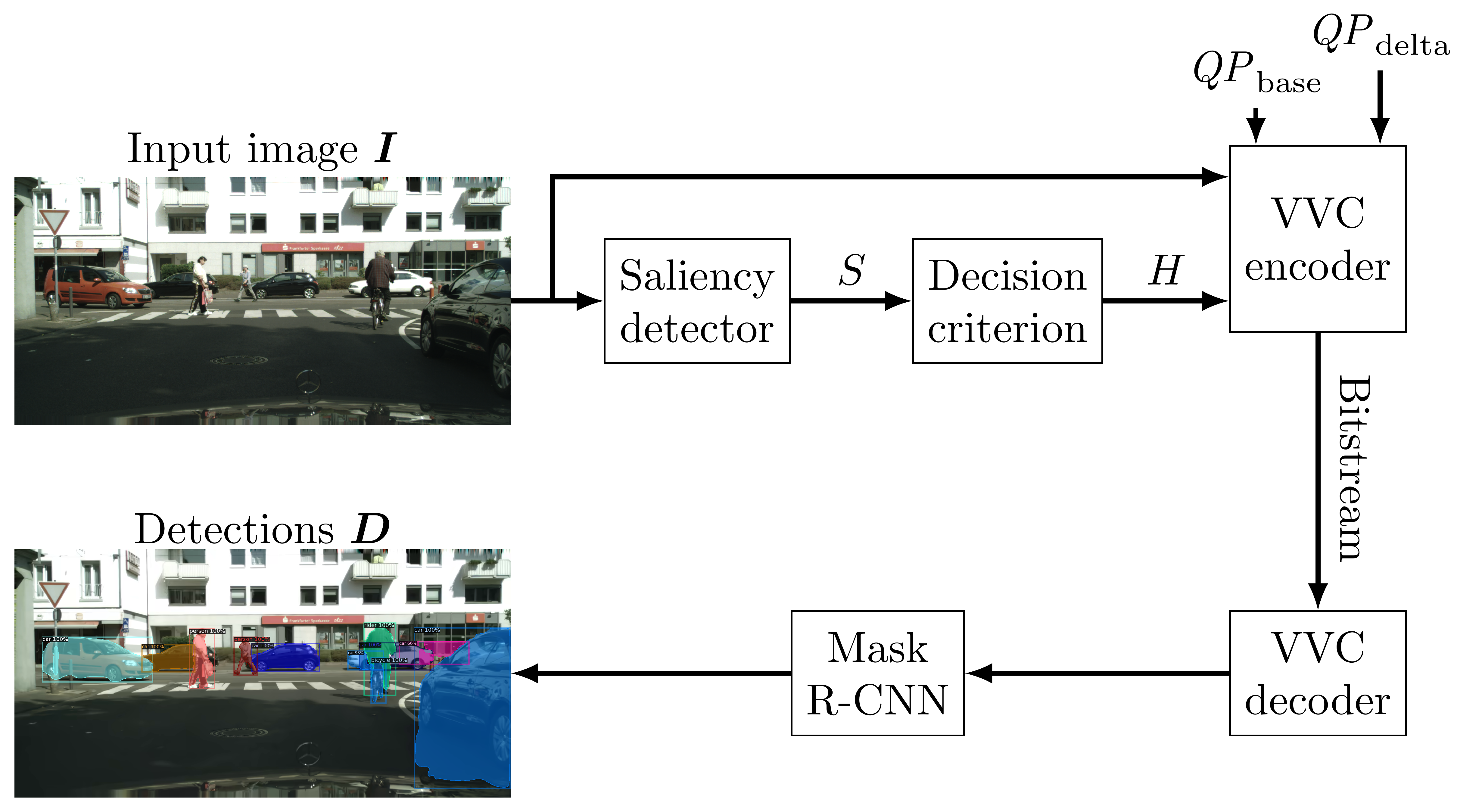}
	\caption{VCM framework with adaptive QP.}
	\label{fig:vcm framework}
	\vspace{-6mm}
\end{figure}

Because of this increasing demand in compression for machines, we investigated whether such an adaptive QP coding chain for VVC as proposed in~\cite{helmrich2019} can also be employed to increase the coding efficiency, when the decoded frame is finally analyzed by a machine rather than being observed by a human. The general setup of the proposed coding framework is shown in Fig.~\ref{fig:vcm framework}. There, we optimize the encoder chain for the state-of-the-art segmentation network Mask R-CNN~\cite{he2017} being applied at the decoder side. In order to significantly reduce the required bitrate and still preserve the accuracy, we adjust the QPA from~\cite{helmrich2019} so that the encoder can focus on the areas containing objects that are important for image segmentation by Mask R-CNN. For that, the selection of the algorithm to detect these salient regions before encoding is vital. On the one hand, it has to precisely find the important regions containing relevant objects, but on the other hand, the saliency detector still has to be fast enough for real-time applications. Thus, we select the one shot-object detection network You Only Look Once~(YOLO)~\cite{redmon2016_ieee} to classify the important regions for encoding. 

To the best of our knowledge, this is the first time that such a framework has been investigated with VVC and Mask R-CNN as evaluation network. In addition, we propose a decision criterion which derives when a CTU has to be considered as salient from the given saliency detections. Besides, also other possible saliency detectors than YOLO are tested.


In the recent past, saliency-driven coding frameworks for a human observer have already been investigated for Joint Photographic Experts Group~(JPEG) and High-Efficiency Video Coding~(HEVC) in~\cite{hosu2016} and~\cite{wei_2016}, respectively.
Using saliency detection algorithms to improve the coding performance for the VCM task has also been researched for older coding standards than VVC.
The approach from Galteri et al.~\cite{galteri2018} creates a binary saliency map by a support vector machine.
From that saliency map, the QP in HEVC is increased in CTUs that can be considered as background. Lastly, the approach in~\cite{choi2018} allocates more bits in HEVC to areas that have been determined being important for the detection task by creating an importance map from the first few layers of the used object detection network.

\section{Proposed Saliency Coding Framework}
\label{sec:proposed saliency coding framework}

\def\imSize{4cm}
\begin{figure}[t]{}
	\centering
	
	\begin{subfigure}[t]{\imSize}
		\centering
		\includegraphics[width=\textwidth]{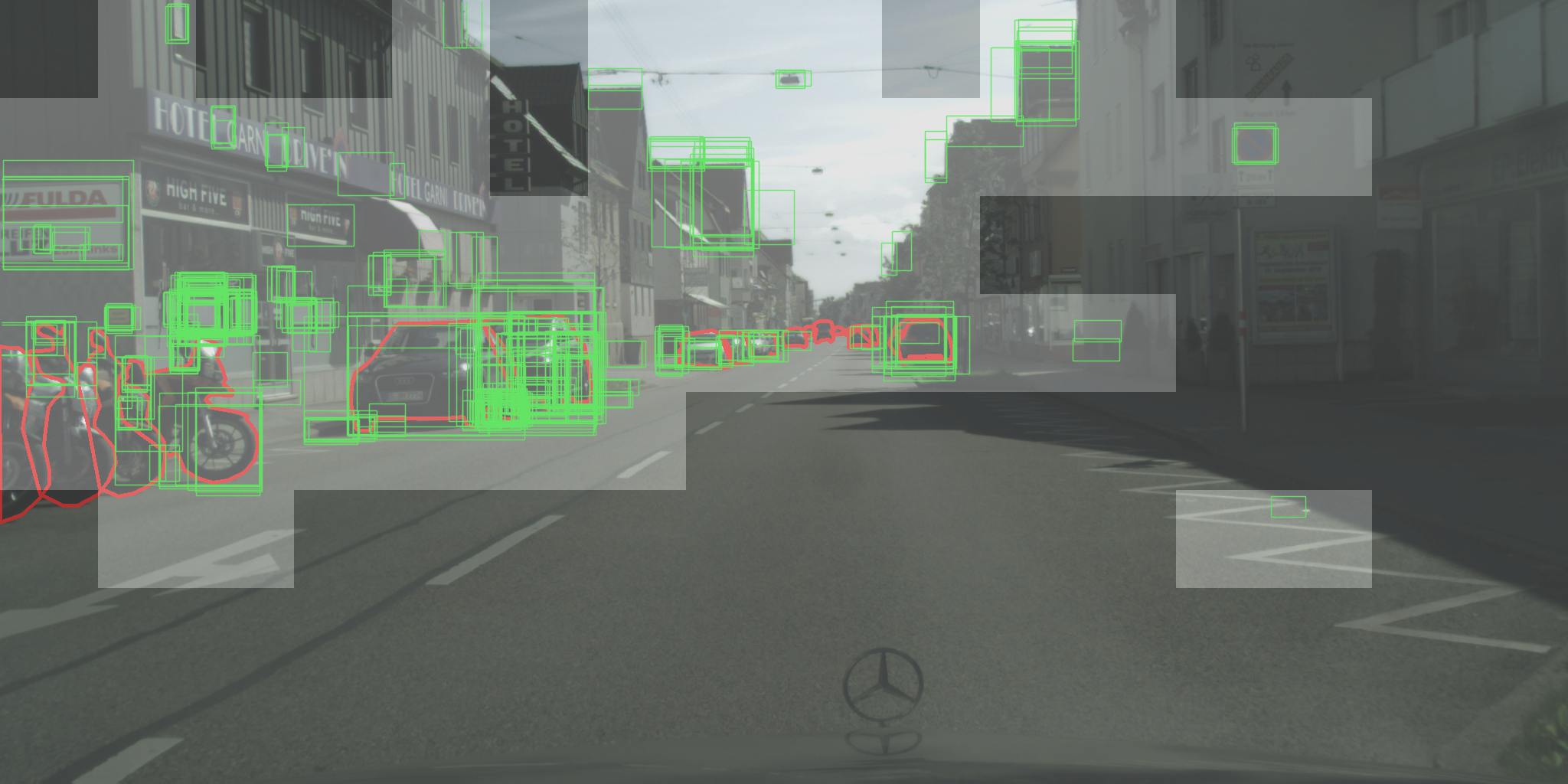}
		\subcaption{Edge Boxes}
	\end{subfigure}
	\hfil
	\begin{subfigure}[t]{\imSize}
		\centering
		\includegraphics[width=\textwidth]{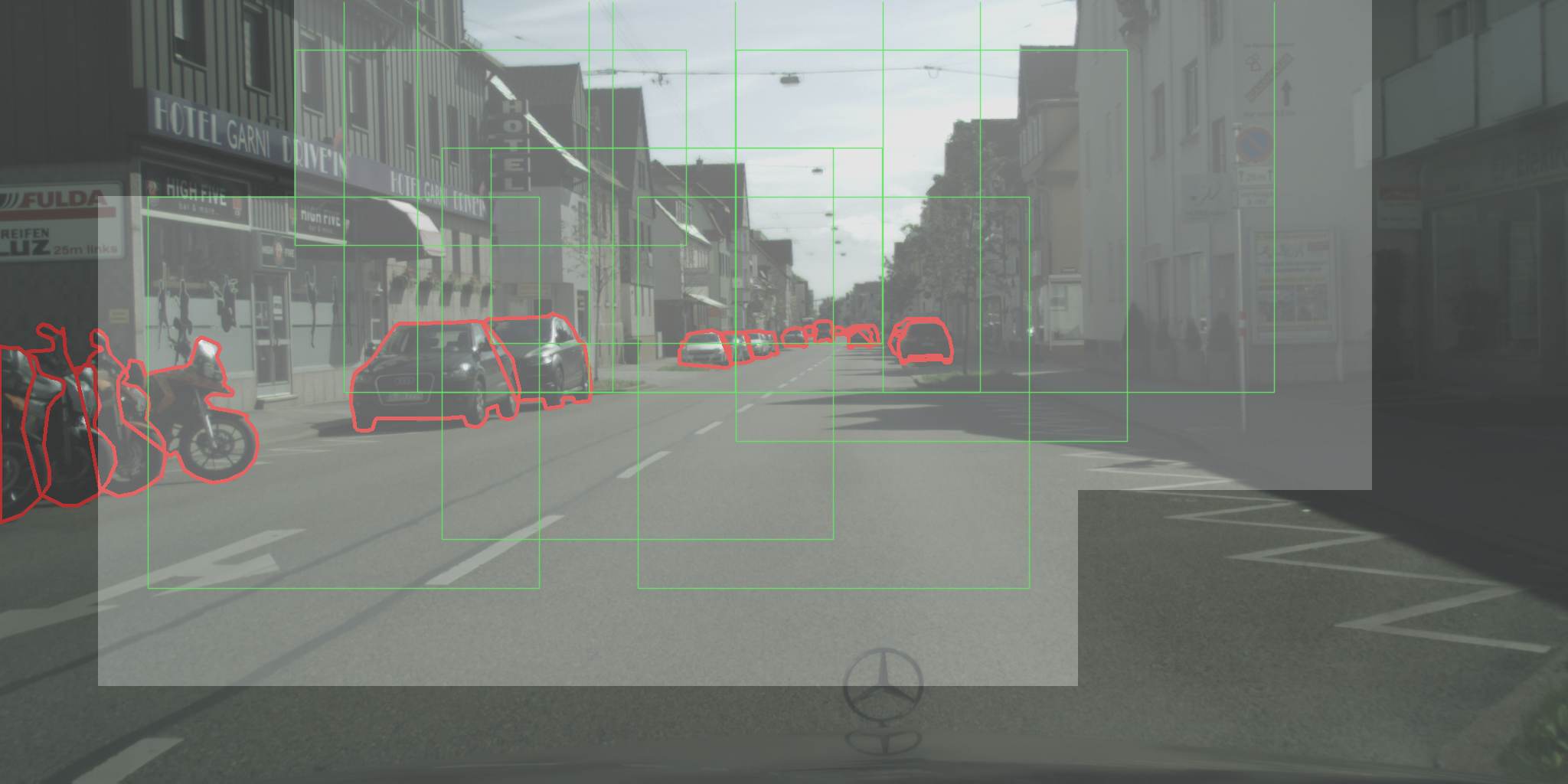}
		\subcaption{BING}
	\end{subfigure}

	\begin{subfigure}[t]{\imSize}
		\centering
		\includegraphics[width=\textwidth]{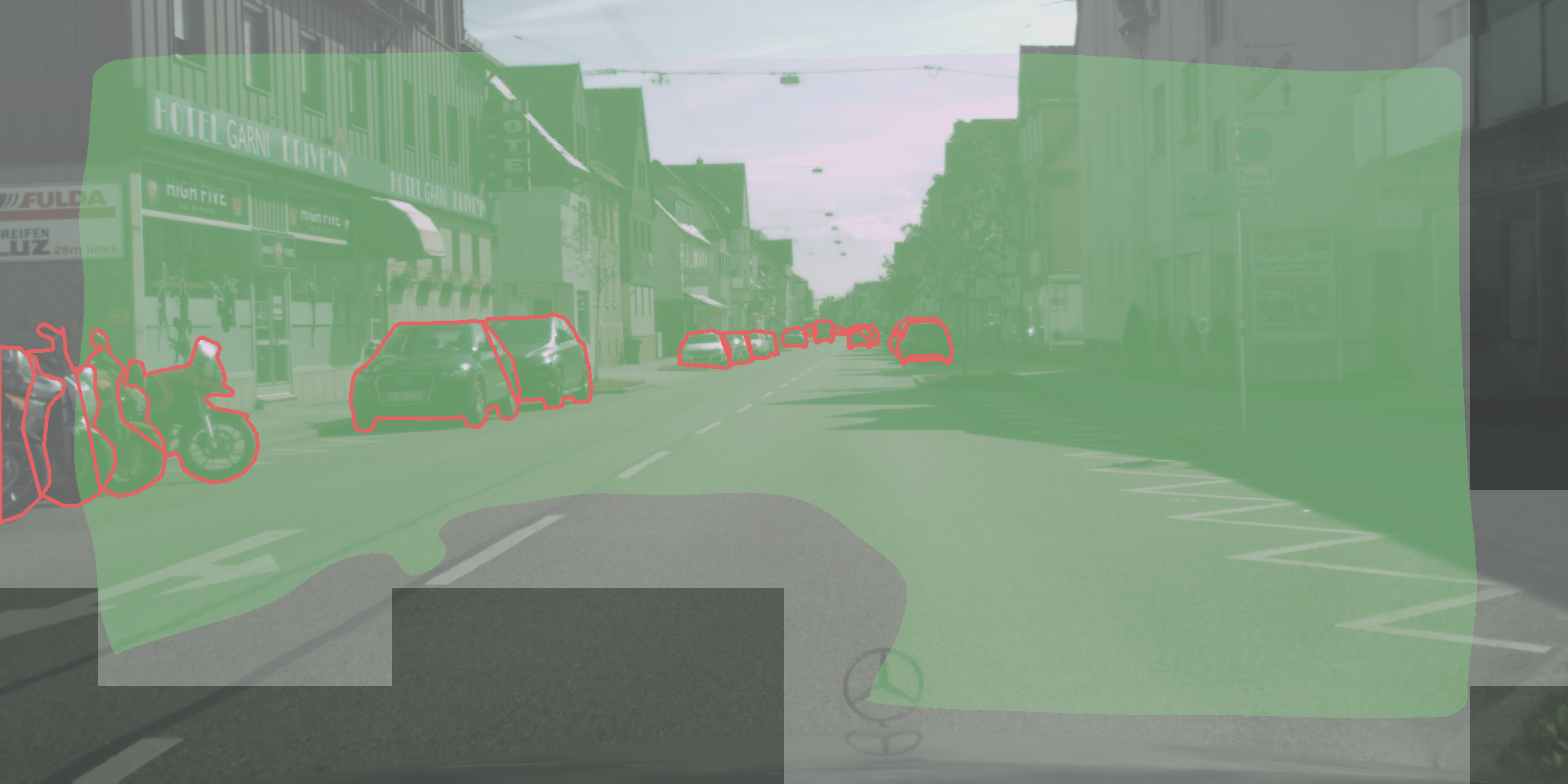}
		\vspace{-3mm}
		\subcaption{FES}
	\end{subfigure}
	\hfil
	\begin{subfigure}[t]{\imSize}
		\centering
		\includegraphics[width=\textwidth]{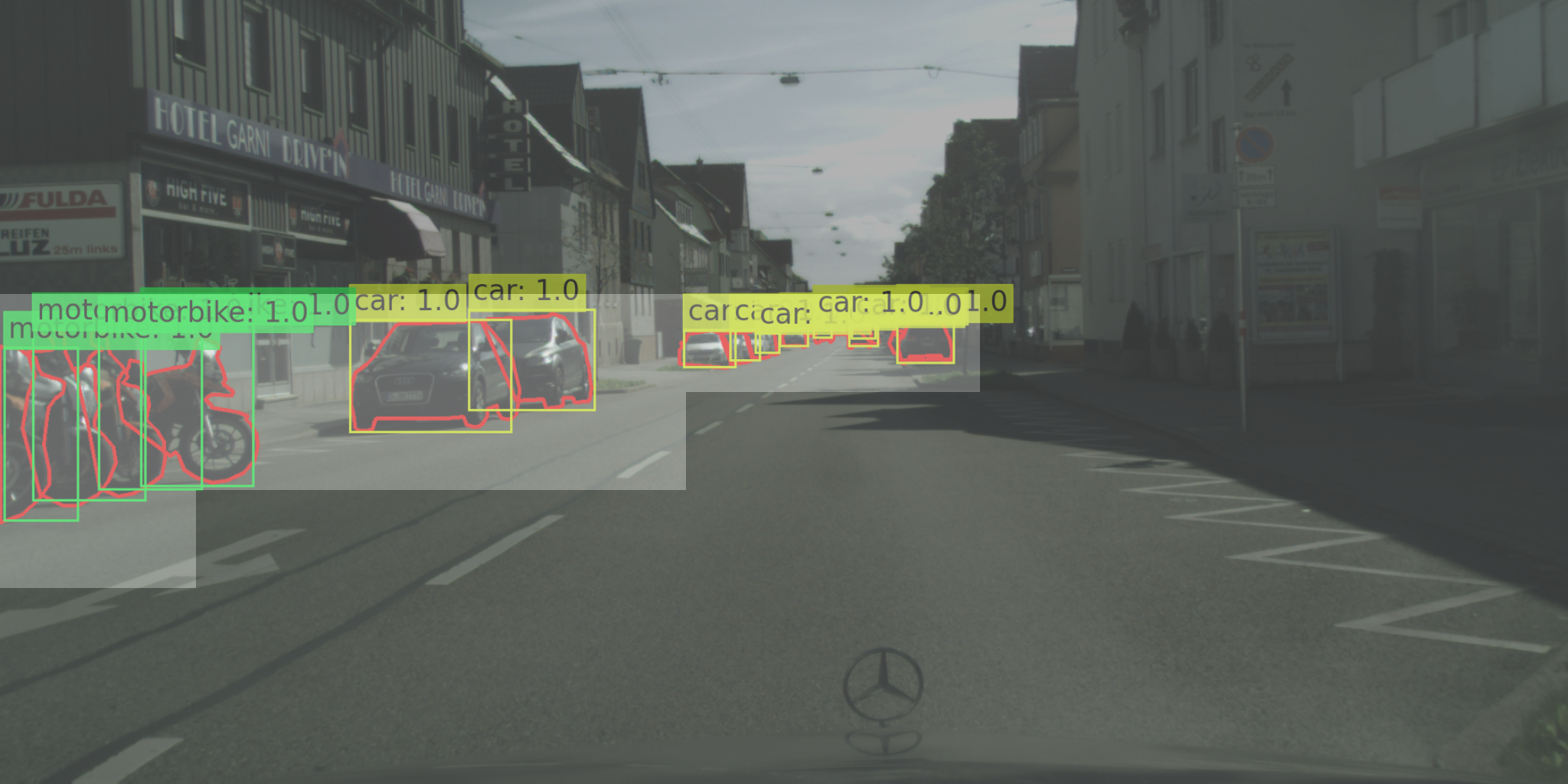}
		\vspace{-3mm}
		\subcaption{YOLO}
	\end{subfigure}
	\vspace{-3mm}
	\caption{Comparison of the different saliency detectors obtaining the salient CTUs exemplary for the Cityscapes image \textit{stuttgart\_000003\_000019\_leftImg8bit.png}. The boxes, or green pixels for FES, show the areas classified as salient and the red edging shows the ground-truth instances. The brighter areas represent the corresponding CTUs that have been classified as salient.}
	\label{fig:saliency algorithms}
	\vspace{-6mm}
\end{figure}

The proposed saliency coding framework mainly consists of two parts before encoding the input image $\inputImage$ with VVC. First, $\inputImage$ is fed into a saliency detector returning the detected objects $\detectionsYOLO$ that span the area of all salient regions that have to be encoded in the best-possible quality. From these detections, a decision criterion is applied to define the quality with which each CTU of size $128\times128$ pixels is encoded.

\subsection{Traditional Saliency Detectors for VCM Task}
\label{traditional saliency detectors for VCM task}
The proposed saliency coding framework can be run with any saliency algorithm returning salient regions or bounding box proposals of possible objects. To that end, three traditional methods are investigated that can be used to classify salient regions for a VCM scenario and which are fast enough to be applied to the encoding process in a practical scenario. The first method is called Edge Boxes~\cite{zitnick2014_edgeboxes} and creates bounding box proposals of objects without classification, where a user-defined number of boxes is returned. Another algorithm returning bounding boxes of objects based on a feature representation of normed gradients is Binarized Normed Gradients~(BING)~\cite{cheng2014_bing}. Furthermore, we consider a classic saliency algorithm for human visual system called Fast and Efficient Saliency~(FES)~\cite{tavakoli2011}, which is based on a Bayesian model.

\subsection{Applying YOLO as Saliency Detector}
\label{subsec: Applying YOLO as saliency detector}

In YOLO, several objects with their bounding boxes are proposed by a convolutional neural network. After refining the boxes, only the most probable objects are taken according to a non-maximum suppression~(NMS) threshold. Only proposals above the threshold are considered as detections. As parametrization for YOLO, we select a NMS threshold of 0.1, and an intersection over union threshold of 0.5, which makes sure that if detections overlay too much with each other, only the one with the highest confidence is considered as detection. These two parameters are differently selected compared to the default parametrization in order to provide a high recall finding the salient CTUs. An example of applying the three traditional saliency algorithms and YOLO to an image is provided in Fig.~\ref{fig:saliency algorithms}.

\subsection{Decision Criterion}
\label{subsec: decision criterion}
For the proposed processing chain, it is vital to select a proper decision criterion deriving the salient CTUs from the detections $\detectionsYOLO$. For that purpose, the overlap area $A^\mathrm{overlap}_{i,k}$ between the current CTU $A^\mathrm{CTU}_{k}$ and the bounding box of a single saliency detection $A^\mathrm{det}_{i}$ is calculated by
\begin{equation}
	A^\mathrm{overlap}_{i,k} = A^\mathrm{CTU}_{k} \cap A^\mathrm{det}_{i},
\end{equation}
with $i$ and $k$ representing the detection index and CTU index, respectively.

To find a suitable threshold when to define a CTU as salient, two cases have to be considered. In the first case, a single detected object is smaller than the size of a CTU. Thus, $A^\mathrm{overlap}_{i,k}$ cannot be larger than the size of the detection $A^\mathrm{det}_{i}$. In the second case, the detection is larger than the size of a CTU, and thus, $A^\mathrm{overlap}_{i,k}$ cannot be larger than the size of the CTU $A^\mathrm{CTU}_{k}$. However, a CTU could be considered as salient in both cases for the subsequent encoding process. Thus, we apply the following criterion that calculates the relative overlap $d$ related to the smaller area of either the detection or a CTU to satisfy both cases by
\begin{equation}
	d_{i,k} = \frac{A^\mathrm{overlap}_{i,k}}{\min(A^\mathrm{CTU}_{k}, A^\mathrm{det}_{i})}.
\end{equation}
For both cases, where either the detection is completely located in a CTU or a CTU is completely covered by the detection, $d$ equals to a maximum value of 1. When the bounding box of the detection and the CTU do not overlay, $d$ equals 0. With this measure, $k$-th CTU is considered as a salient CTU, when at least one detection has a relative overlap with the CTU larger than a threshold $\theta$ by 
\begin{equation}
\salientCTUs_k = \left\{
\begin{array}{ll}
\ 1 & \textrm{if} \quad \displaystyle\max_{i,k} (d_{i,k}) > \theta \\\
0 & \textrm{else}\\
\end{array}
\right. ,
\label{eq:decision device}
\end{equation}
with $\salientCTUs$ saving the saliency information for each CTU. The optimal choice of $\theta$ will be evaluated later in Section~\ref{subsec: Determining the optimal parametrization}.

%

In the proposed processing chain, the VVC encoder requires the input image $\inputImage$ and the decision mask $\salientCTUs$. Besides, the encoder is steered with the user-defined base and delta QP values $\QP_\mathrm{base}$ and $\QP_\mathrm{delta}$, respectively. With these, the adapted $QP_k$ for $k$-th CTU can be calculated according to

\begin{equation}
\QP_k = \left\{
\begin{array}{ll}
\QP_\mathrm{base} & \salientCTUs_k = 1  \quad \textrm{(salient)}\\
\QP_\mathrm{base} + \QP_\mathrm{delta} & \salientCTUs_k = 0 \quad \textrm{(non-salient)}\\
\end{array}
\right. .
\label{eq:decision device 2}
\end{equation}
Subsequently, the resulting bitstream gets decoded and the decoded frame is taken as input for Mask R-CNN to detect and segment the objects $\detections$, which each consist of a class label, a segmentation mask, and a detection certainty. Eventually, the detection accuracy of Mask R-CNN and the required bitrate are measured, to evaluate the coding quality of the proposed VVC framework.

\section{Analytical Methods}

\subsection{Dataset}

In order to evaluate the proposed coding framework, we selected the Cityscapes dataset~\cite{cordts2016} containing single images capturing road scenes with a resolution of $2048\times 1024$ pixels. The dataset includes uncompressed images and their corresponding pixel-wise annotated ground-truth data for eight classes of road-users: bicycle, bus, car, motorcycle, person, rider, train, and truck.

To quantify the detection accuracy of Mask R-CNN, analyzing the decoded and deteriorated frame, the average precision~(AP) metric calculated for pixel-accurate detections was taken as described in~\cite{cordts2016}, which is calculated for each class. 
Instead of averaging the AP values equally over all classes as done in~\cite{cordts2016}, we weight the AP values of each class according to the number of instances as proposed in~\cite{fischer2020_ICIP}, since some classes are underrepresented in the Cityscapes validation set, e.g. bus and truck.

\subsection{Setup}

To investigate the coding gains of the proposed coding framework we used the 500 uncompressed Cityscapes validation images as input data. These were first analyzed by YOLO to obtain the detections. From these detections, the salient CTUs were derived according to \eqref{eq:decision device} and \eqref{eq:decision device 2}. Afterwards, the 500 images were encoded with VTM-6.2~\cite{chen2019vtm6} and $\QP_\mathrm{base}$ values of 12, 17, 22, and 27 in all intra configuration. Subsequently, the decoded images were fed into Mask R-CNN to obtain the segmented objects and measure its performance on deteriorated images with the weighted AP metric. The Mask R-CNN code was taken from~\cite{wu2019detectron2} as well as the model that has already been trained on the Cityscapes training dataset. In order to measure the coding gains for the VCM task, we calculated the Bj\o ntegaard delta rate (BDR)~\cite{bjontegaard2001_new} and substituted the commonly used quality metric PSNR used with the weighted AP values, as practiced in~\cite{choi2018}, \cite{fischer2020_ICIP} and \cite{fischer2020_FRDO} and as suggested by the JVET VCM group~\cite{liu2020_VCM_CTC}.

For YOLO, we used the \textit{darknet} YOLO-v3 implementation from~\cite{redmon_yolo_darknet_git} and their model that has already been trained on the Microsoft COCO dataset~\cite{lin2014_COCO}. Since COCO covers more instance classes than Cityscapes, all found objects by YOLO that are not related to a Cityscapes class have not been considered for the saliency detection. The code for the traditional saliency algorithms Edge Boxes and BING was taken from \textit{OpenCV}~\cite{opencv_contrib} and for FES we used the reference code from~\cite{tavakoli_fes_git}.

\vspace{2mm}
\section{Experimental Results}
\label{sec:experimental results}

\subsection{Determining the Optimal Parametrization}
\label{subsec: Determining the optimal parametrization}

In general, the coding quality of the proposed coding framework mainly depends on two parameters defined in Section~\ref{subsec: decision criterion}. First, the threshold $\theta$ in the decision criterion defines the minimum relative overlap between a detected salient object and a CTU to classify this CTU as salient. With a higher $\theta$, more CTUs covering only small boundary areas of detections are encoded with low quality, which might not harm the Mask R-CNN accuracy, but results in less required bitrate. Contrary, even encoding those small areas with less quality might lead to missed detections at the decoder side. Similar considerations have to be made for choosing the QP offset $\QP_\mathrm{delta}$ for non-salient CTUs. With a high $\QP_\mathrm{delta}$ the bitrate is reduced more, with the drawback of missed detections at the decoder side because of CTUs being mistakenly classified as non-salient before encoding.

\begin{table}[]
	\centering
	\normalsize
	\caption{BDR values in \% with the weighted AP as quality metric for $\QP_\mathrm{base} \in \{12,17,22,27\}$ with VTM-6.2 as anchor for different $\theta$ and $\QP_\mathrm{delta}$ values with YOLO as saliency detector. Bold values indicate the highest coding gains for each $\QP_\mathrm{delta}$.}
	\begin{tabular}{cllll}
		\hline
		& \multicolumn{4}{c}{$\theta$}     \\ 
		$\QP_\mathrm{delta}$ & 0.0   & 0.025 & 0.05  & 0.1   \\ \hline
		5  & -14.8 & -15.2 & -15.9 & \textbf{-16.4} \\
		20 & -26.0 & \textbf{-27.3} & -26.3 & -26.1 \\
		max & \textbf{-29.0} & -26.8 & -28.4 & -26.4 \\ \hline
	\end{tabular}
	\label{tab:BDR values theta and delta QP}
	\vspace{-3mm}
\end{table}

For $\theta$, we tested values of 0.0, 0.025, 0.05, and 0.1 to receive the salient CTUs. As delta QP for the non-salient CTUs, we tested values of 5, 20, and the case of setting it to the maximum possible QP of 63. The resulting BDR-values with the standard VTM-6.2 without QPA as anchor are listed in Table~\ref{tab:BDR values theta and delta QP}.

All investigated parameter combinations of $\theta$ and $\QP_\mathrm{delta}$ provide large coding gains for the proposed processing chain over standard encoding with VTM-6.2. The highest coding gains can be achieved, when the QP is set to its maximum value of 63 for non-salient CTUs, resulting in a bitrate reduction of 29.0\,\% for the same accuracy of Mask R-CNN at the decoder side. For this case, setting the threshold $\theta$ of the decision criterion to 0.0 results in the best coding performance. From these results, it can be derived that first, each CTU containing YOLO detections should be regarded as salient and second, non-salient CTUs can be coded with the lowest possible quality without degrading the detection accuracy of Mask-RCNN.

%

\subsection{Comparison of Different Saliency Algorithms}
\label{subsec: Comaprison of different saliency algorithms}

In this section, the proposed coding framework with YOLO employed as saliency detection algorithm is compared against the other saliency detection algorithms presented in Section~\ref{subsec: Applying YOLO as saliency detector}.

\begin{figure}[t]{}
	\centering
	\includegraphics[width=0.39\textwidth]{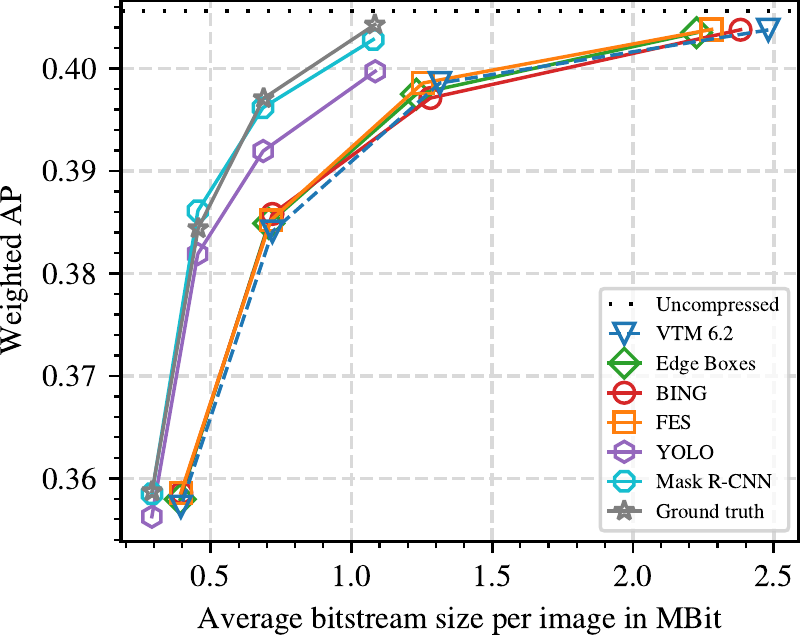}
	\caption{Weighted AP values over bitrate comparing the different saliency detectors with the reference VTM-6.2 for $\QP_\mathrm{base} \in \{12,17,22,27\}$, $\theta=0$ , and $\QP_\mathrm{delta}$ set to maximum. The dotted black line represents the weighted AP on uncompressed input images.}
	\label{fig:weighted ap over bitrate}
	\vspace{-6mm}
\end{figure}

In addition to these algorithms, the framework was also evaluated with taking the Mask R-CNN detections before encoding as saliency information, as well as the ground-truth data. This was supposed to serve as an oracle test assuming a nearly perfect saliency detector to find out the maximum possible coding gains of the investigated processing framework. For these two oracle tests as well as for the three other saliency algorithms, the same decision criterion was applied as for YOLO with $\theta=0$ and $\QP_\mathrm{delta}$ set to the maximum, since this resulted in the highest coding gains. The resulting weighted AP over rate curves are shown in Fig.~\ref{fig:weighted ap over bitrate}.

YOLO as saliency detector is able to significantly reduce the required bitrate compared to the standard VTM. These savings are even as high as for both oracle tests of using Mask R-CNN and the ground-truth data to determine the salient CTUs. However, the resulting weighted AP is not as high, which is caused by the fact that YOLO is not as accurate detecting all salient objects as Mask R-CNN and the ground-truth. The other three algorithms Edge Boxes, BING, and FES are able to increase the coding performance as well. But, they are not able to reach the coding gains of using YOLO as saliency detector. The corresponding BDR values are listed in Table~\ref{tab:BDR values high recall} for multiple values of $\QP_\mathrm{delta}$. These numbers confirm that setting QP to the maximum value results in the highest bitrate savings when the detection of salient CTUs is accurate enough, which is true for YOLO and the oracle test. Coding gains from the traditional saliency detectors (always below 7\,\%) indicate that YOLO is better adapted for this application. In Fig.~\ref{fig:saliency algorithms} the reason for this observation can be spotted. YOLO is able to find all significant CTUs while still being precise enough to not falsely classify non-salient CTUs as salient, which would result in less coding efficiency.

Furthermore, YOLO takes around 0.6 seconds for one Cityscapes image on an Intel Xeon\textsuperscript{\textcopyright} E3-1275 CPU and is thus still applicable as saliency detector before encoding in terms of runtime for practical applications, without requiring a dedicated GPU. For comparison, Mask R-CNN takes around 7.5 seconds on the same CPU.


%

\begin{table}[]
	\centering
	\caption{BDR values in \% with the weighted AP as quality metric for $\QP_\mathrm{base} \in \{12,17,22,27\}$ and $\theta=0$ with VTM-6.2 as anchor for the different delta $\QP_\mathrm{delta}$ values in non-salient regions. Bold values indicate the highest coding gains for a given saliency detection algorithm.}
	\normalsize
	\begin{tabular}{lllll}
		\hline
		& \multicolumn{4}{c}{$\QP_\mathrm{delta}$}                                            \\
		Saliency Detector & \multicolumn{1}{c}{5} & \multicolumn{1}{c}{10} & \multicolumn{1}{c}{20}& \multicolumn{1}{c}{max} \\ \hline
		Edge Boxes         & \textbf{-5.9}        & -5.4                  & -3.9                  & -2.7\\
		BING         & \textbf{-6.1}        & -0.8                  & -4.0                  & -2.9 \\
		FES                & \textbf{-5.0}        & -2.11                  & -1.3            & -3.9      \\
		YOLO               & -11.4                & -22.1                 & -26.0  & \textbf{-29.0}      \\
		Mask R-CNN         & -18.7                & -24.9                 & -32.4  & \textbf{-37.3}      \\
		Ground truth       & -18.5               & -27.2                 & -33.5  & \textbf{-36.8}      \\ \hline
	\end{tabular}
	\label{tab:BDR values high recall}
	\vspace{-3mm}
\end{table}

\vspace{-3mm}
\section{Conclusion}
In this paper, we investigated an image coding framework with VVC and adaptive QP for VCM scenarios and the real-time-capable object detection network YOLO to obtain the salient CTUs. We found that for the best parametrization, using YOLO results in bitrate savings of 29\,\%, where maximum savings of around 37\,\% could be obtained using oracle tests. In future work, one might investigate to adapt the delta QP according to the certainty of the saliency detection network. Besides, the saliency information can be used in the rate-distortion optimization of the encoder to further adapt VVC for the VCM task.

\bibliographystyle{IEEEtran}
\setstretch{0.95}
\bibliography{/home/fischer/Paper/jabref_literature_research_ms2.bib,/home/fischer/Paper/literature_M2M_communication.bib,/home/fischer/Paper/jabref_used_software.bib}

\end{document}